\documentclass{article}

\usepackage{microtype}
\usepackage{graphicx}
\usepackage{booktabs} 
\usepackage{subfig}

\usepackage{hyperref}
\usepackage{mathtools}

\usepackage[accepted]{icml2020}

\icmltitlerunning{Image Set Visual Question Answering}

\begin{document}

\twocolumn[
\icmltitle{Analysis on Image Set Visual Question Answering}



\icmlsetsymbol{equal}{*}

\begin{icmlauthorlist}
\icmlauthor{Abhinav Khattar}{}
\icmlauthor{Aviral Joshi}{}
\icmlauthor{Har Simrat Singh}{}
\icmlauthor{Pulkit Goel}{}
\icmlauthor{Rohit Prakash Barnwal}{}
\end{icmlauthorlist}



\icmlkeywords{Machine Learning, ICML}

\vskip 0.3in
]




\begin{abstract}

We tackle the challenge of Visual Question Answering in multi-image setting for the ISVQA dataset. Traditional VQA tasks have focused on a single-image setting where the target answer is generated from a single image. Image set VQA, however, comprises of a set of images and requires finding connection between images, relate the objects across images based on these connections and generate a unified answer. In this report, we work with 4 approaches in a bid to improve the performance on the task. We analyse and compare our results with three baseline models - LXMERT, HME-VideoQA and VisualBERT - and show that our approaches can provide a slight improvement over the baselines. In specific, we try to improve on the spatial awareness of the model and help the model identify color using enhanced pre-training, reduce language dependence using adversarial regularization, and improve counting using regression loss and graph based deduplication.  We further delve into an in-depth analysis on the language bias in the ISVQA dataset and show how models trained on ISVQA implicitly learn to associate language more strongly with the final answer.
\end{abstract}

\section{Introduction}
\label{sec:intro}

\begin{figure}[!htpb]
    \centering
    \includegraphics[width=\columnwidth]{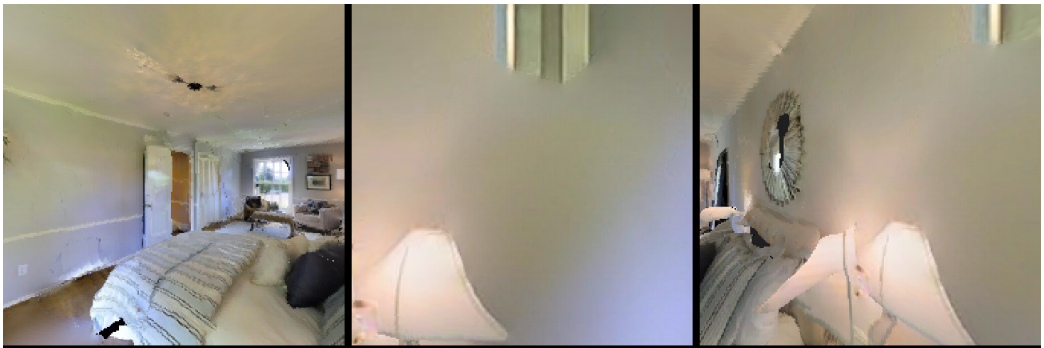}
    \caption{Image from the ISVQA dataset. The associated question with the image is, "$\textit{What is hanging above the bed?}$" and with the corresponding answer is, "$\textit{mirror}$".}
    \label{fig:intro}
\end{figure}

Image Set Visual Question Answering (ISVQA) \cite{bansal2020visual} is the task of answering a question given a set of Images. In this paper we explore the task of ISVQA, identify relevant research challenges and propose well defined strategies inspired by error analysis and thorough investigation of Visual Question Answering \cite{antol2015vqa} a closely related task. Tasks that combine vision and natural language continue to inspire considerable research at the boundary of computer vision and natural language processing. Since its introduction, VQA has attracted significant attention from the research community as answering natural language questions about images requires understanding a wide range of detailed semantics of an image and how they are referred to in natural language. Furthermore, solving VQA is of practical importance given its utility in assisting visually impaired people. While the problem of ISVQA is quite similar to that of VQA, the former can be significantly more challenging than vanilla VQA since the answer is not necessarily contained within a single image and thus reasoning over objects and concepts in different images becomes paramount for achieving good performance on ISVQA. For example, In Fig~\ref{fig:intro}, a model has to find the relationship between the bed in the left image and the mirror in the right, via pillows which are common to both the images.

ISVQA reflects information retrieval from multiple images of relevance but with no obvious continuous correspondences. A model for solving this type of problems has to understand the question, find the connections between different images, use these connections to relate objects across images and generate the answer. ISVQA leads to new research challenges, including: a) How to use natural language to guide scene understanding across multiple views/images. b) How to fuse information from relevant images to reason about relationships among entities. These challenges associated with scene understanding do not exist in single-image VQA settings but frequently happen in the real world. While ISVQA task includes answering questions about images taken at different times (e.g. like in camera trap photography, or images under same story on Facebook/Instagram), images taken at different locations (e.g. security cameras capturing different locations within a store, or pictures of same house on real estate websites), or images taken from different viewpoints (e.g. live sports coverage), we focus on images from multiple views of an environment.

In this paper, we explore the specific challenges and issues faced while solving the problem proposed in ISVQA. To the best of our knowledge, this is the first work on solving the problem of Image Set Visual Question Answering, apart from the initial ISVQA dataset paper. Initially, we study the LXMERT baseline for ISVQA and discuss where the model lacks. Based on our findings, we focus on the issues of counting, color awareness, spatial awareness, and language dependence, proposing approaches to tackle each of these issues and finally improving the performance over the LXMERT baseline. We also discuss in-depth the problem of language bias in ISVQA, show how it significantly affects a model being trained on the dataset, and demonstrate how our adversarial regularization strategy stops the model from being overconfident in its prediction in the absence of visual modality. We finally call for work on a VQA-CP~\cite{agrawal2018don} like dataset for the problem of ISVQA that helps evaluate how ISVQA models will perform in ``real-life" setting.

In the following section \ref{sec:related}, we present the works that closely relate to ISVQA. We then analyze the ISVQA dataset and evaluation metrics in section \ref{sec:setup}. We also formulate the problem statement and introduce the Multimodal baselines for ISVQA dataset in this section. Section \ref{sec:proposed-approach} discusses the 4 research ideas that we propose in this paper; namely the adversarial regularization, enhanced VQA, count-aware ISVQA and Regression loss based improvements. In section \ref{sec:results} we present results obtained on the 4 proposed models and present an in-depth anaysis of adversarial regularization. Finally, we conclude and discuss future work in section \ref{sec:conclusion}.

\section{Related Work}
\label{sec:related}

Being a relatively new task, the only available work on ISVQA is the dataset paper. The paper describes 4 baseline architectures and concludes that their Transformer-based architecture - which derives from LXMERT~\cite{tan2019lxmert} - performs the best. The problem of ISVQA lies somewhere at the intersection of VideoQA and VQA where instead of having a single image (as in VQA) we have multiple images (which is the case with VideoQA), but the images do not form a continuous scene (as is the case with VideoQA). Given the similarity of our problem with Visual Question Answering (VQA) and Video Question Answering (VideoQA), in this section, we discuss existing works in both these fields and use existing methodologies from both of these as our baselines.  

\subsection{VQA}
VQA is a well-researched task with multiple datasets including VQA 1.0~\cite{antol2015vqa}, VQA 2.0~\cite{goyal2017making}, DAQUAR~\cite{malinowski2014multi}, and COCO-QA~\cite{ren2015exploring}. VQA 2.0 is the largest, and most commonly used dataset for VQA with over 1.1M questions and over 11M ground-truth answers.

While most current research in VQA involves using some sort of masked pre-training objective, for the sake of completeness, we discuss the 2 major types of models that exist to tackle VQA:

\subsubsection{RNN based models without masked pre-training}

\citet{anderson2018bottom} introduced an approach which combines top-down(soft) and bottom-up (hard) attention mechanisms to achieve state-of-the-art performance on image captioning and obtain first place in the VQA 2017 challenge. The proposed approach utilized the Faster-RCNN model to identify objects of interest in images and a ResNet model to extract vector representations of those objects. The answer is generated using a 2 layer LSTM network where the first layer is referred to as top-down attention layer and the second deals with generating the final output for both image captioning and VQA. The top-down LSTM attention  layer involved utilizes a concatenation of word embedding, image embedding and previous time-steps hidden-state of the 2nd layer as input. The hidden-state generated by the top-down attention layer is then fed to the Language LSTM layer which attends to the embeddings of the extracted regions generated by the bottom-up attention model. Finally the output of the Language LSTM layer is used to compute a probability distribution over the vocabulary to generate the caption in the case of Image Captioning and the answer in the case of Visual Question Answering.

\subsubsection{Transformer based models with masked pre-training}
\label{sec:VQApretrain}

Recent works in the field of NLP have brought to light the positive effects of Language Model pre-training on various downstream tasks~\cite{devlin2018bert, peters2018deep, lample2019cross}. Taking inspiration from such architectures, recent VQA models adopt a BERT~\cite{devlin2018bert} style of pre-training for Transformer models~\cite{vaswani2017attention}. Some of these VQA models include: Visual BERT~\cite{li2019visualbert}, VL-BERT~\cite{su2019vl}, UNITER~\cite{chen2019uniter}, and OSCAR~\cite{li2020oscar}. Using a Transformer with a pre-training objective is a common theme that runs across all of these models. 

VisualBERT, which is probably the most simplistic architecture out of the three techniques mentioned, was one of the first VQA architectures to come out with a BERT-like architecture. In an attempt to learn implicit alignments between text and images, the model is trained on 2 pre-training objectives - (1) Masked Language Objective, (2) Sentence Image prediction. While the first objective is completely identical to BERT, the model, in this case, takes as input both the text and the image regions. The second objective, on the other hand, is unique and attempts to answer if the text provided with the image is the image’s true caption. The authors of VisualBERT first pre-train the model on MS-COCO, followed by pre-training on the VQA dataset, finally followed by task-specific fine-tuning.

Works using Transformers that came after VisualBERT focus on how to learn better alignments and relationships between text and images. VL-BERT throws away the Sentence Image prediction brought in by VisualBERT and add a Masked Region of Interest Classification task which masks out Region of Interests (RoI) and then asks the model to predict the entity that may have been masked out. This objective helps learn stronger alignments and as a result, helps VL-BERT outperform VisualBERT. To further improve over VL-BERT, UNITER incorporates Image-Text Matching, Word-Region Alignment, Masked Region Feature Regression, Masked Region Classification, and Masked Region Classification with KL-Divergence objectives. OSCAR - which is the current SOTA on VQA 2.0 - outperforms UNITER by further helping image-text alignment using object-tags detected from images while just using Masked Token and Contrastive Loss. Another recent model, LXMERT~\cite{tan2019lxmert}, diverges a bit from the standard Transformer-Encoder architecture and uses an attention based architecture with different initial model segments for both the modalities. The architecture is also pre-trained on various pre-training objectives including Masked LM, Masked Object Precition, Cross-Modality Matching, and Image Question Answering. We use this LXMERT architecture as on of the baselines for ISVQA task as it is easy to directly use this model for our task. 

Owing to the strong positive effects of transfer learning and pre-training, recent Transformer-based pretrained models outperform the non pretrained RNN-based ones. The Transformer models, though, use the same visual features as the ones used by RNN models which are most commonly extracted using an R-CNN based feature extractor. As a whole, the modern Transformer-based architectures provide a more expressive and elegant solution to the problem of VQA learning from a glut of other information sources using an array of objectives that help the model learn better and more generalizable representations.

\subsection{VideoQA}
Compared to image-based VQA, there has been less work done on video-based VQA. In past few years, several video QA datasets have been proposed, e.g. MovieFIB~\cite{chen2017movie}, MovieQA~\cite{tapaswi2016movieqa}, TGIF-QA~\cite{jang2017tgif}, MarioQA~\cite{mun2017marioqa} and TVQA~\cite{lei2018tvqa}. TVQA is one of the largest video QA datasets, providing a large video QA dataset built on top of 6 famous TV series containing 152.5K multiple choice questions from 21.8K, 60-90 second long video clips. The image sets in ISVQA dataset are not akin to video frames which are temporally continuous. Also, unlike embodied QA~\cite{das2018embodied}, ISVQA does not have an agent interacting with the environment.

\citet{fan2019heterogeneous} propose a novel Multimodal Fusion based architecture to solve the task of Video QA. The architecture takes as input 2 sets of visual features as appearance features extracted frame by frame using a standard CNN based architecture such as ResNet\cite{he2015deep} or VGG\cite{simonyan2015deep} and motion features extracted from clips of the video using the C3D\cite{tran2015learning} model. GloVe 300-D\cite{pennington2014glove} embeddings based features extracted from an LSTM network form the inputs of the textual modality.
The architecture incorporates an external heterogeneous memory to integrate the two visual features and learn the joint attention. A similar external memory is also implemented for the question features. This helps to understand the global as well as the local context and also helps in multimodal fusion of the text and video modalities. 
The multimodal fusion consists of a core LSTM controller which takes the hidden memory of the combined video and question based features as the input. During each iteration, the controller attends to different parts of the video features and question features with temporal attention mechanism, and combines the attended features. 

Our work focuses on the problem of ISVQA, and to the best of our knowledge, ours is the first work exploring the techniques to solve this problem, apart from the initial dataset paper. As a result, none of our related works try to tackle the problem of ISVQA - adding to the novelty of our work. In addition, some of the techniques that we propose including enhanced pre-training and adversarial regularization are novel and have not been previously explored in other related contexts.


\section{Experimental Setup}
\label{sec:setup}
\subsection{Dataset}
We propose to use the dataset provided by ISVQA. The dataset comprises of 2 modalities - image and language. The authors of ISVQA use existing image set datasets of Gibson \cite{xia2018gibson} and nuScenes \cite{caesar2020nuscenes} as sources to build an ISVQA dataset. Gibson dataset provides 3D indoor scans of buildings, rooms and offices. The scans are of 2 types 1) Gibson Building and 2) Gibson Room. nuScenes contain outdoor scenes generated by a $360^{\circ}$ camera mounted on a car in city streets. The details for the dataset are presented in Table \ref{tab:Table 1}.
\begin{table}[h]
    \centering
    \scriptsize
    \begin{tabular}{l|c|c|c}
         \hline
         Dataset& Train Set& Test Set& Unique Answers  \\ \hline \hline
         Indoor - Gibson& 69,207& 22,272& 961  \\ \hline
         Outdoor - nuScenes& 33,973& 15,644& 650
    \end{tabular}
    \caption{Dataset specifications}
    \label{tab:Table 1}
\end{table}

Currently, we are only working with the NuScenes data as the Gibson data has not entirely been made available by the ISVQA dataset authors.

\subsection{Dataset Analysis}
\label{sec:data-analysis}
Most of the questions phrases contain between 5 and 10 words. Most frequent questions are
about physical properties of objects, and spatial relationships between different entities. The questions can be broadly said to be about the color of objects in the images, relative spatial properties and the count of the objects.

The dataset tries to ensure that there are a significant number of data points which require more than one image to answer a question. This forces the model to learn the entire scene rather than look at just one region of interest. About 7,000 such samples exist in Gibson-Room and about 3,000 in Gibson-Building. In nuScenes only about 1,000 such examples exist but the number of data points which require more than two images is greater than the indoor ones.




\subsection{Evaluation Metric}
Given that the questions are multiple-choice, we propose to use VQA-accuracy as the evaluation metric. In the ISVQA setting, each image set has been annotated by 3 annotators. VQA-accuracy for an ISVQA sample is thus, 1 when it is supported by 2 or more annotators, 0.5 when supported by 1 annotator, and 0 otherwise.  Human performance using standard accuracy measure was calculated in ISVQA. VQA-accuracy of 91.88\% for nuScenes and 88.80\% for Gibson was obtained. This makes the task quite challenging and also provides an interesting observation of why the human performance was not close to 100\%. In many cases, humans gave a similar answer to the question which although conveyed the same meaning, was not semantically similar to the provided "true labels". For example, the answer could be “black and white” but “white and black”, though same would not be considered semantically similar.

\subsection{Problem Statement}
\label{sec:problem}
The problem of ISVQA can thus be formulated as the following. Given an image set $S = [I_1, I_2, \cdots, I_N] $, where $N$ is the total number of images associated with a data point and a question $Q = [q_1, q_2, \cdots, q_T]$, where $q_i$ denotes the $i^{th}$ word in the question of total $T$ words, the model should provide an answer $a$ such that $a = f(S,Q)$. The task is a discriminative question answering task which implies that the model picks the best possible answer from a set of all possible answers. Thus, the function $f$ ideally generates a set of scores or probabilities over all the possible answers in the answer set $A = [a_1, a_2, \cdots, a_M]$. For nuScenes dataset, there are 6 images associated with each data point i.e. $N = 6$ and the total possible answers $M = 650$.

\subsection{Multimodal Basline}
\label{sec:baselines}

Given the lack of work on ISVQA specific baseline models, we implemented and adapted one Video-QA and two VQA models to work in an ISVQA setting. In specific, we tried: HME-Video VQA, VisualBERT, and LXMERT models as our baselines. The HME-Video VQA and LXMERT baselines have been proposed by the authors of the ISVQA dataset, and our re-implementation of these models on ISVQA yields results in the same domain as has been observed by authors of the dataset. 

To adapt HME-VideoQA model to ISVQA, we used the images in the set as the frames of the video so that the appearance features could be calculated directly. We used VGG19 to extract the features. For motion features, we assumed the images in the set to be a part of a 3D video. We used C3D model to extract the features. The visual features were of size 4096. Adam was used as the optimizer with learning rate of $1e-3$. A memory of size $256$ was used for the Heterogeneous memory model. LSTM of hidden memory size $512$ was used in the multimodal fusion module.

For the VQA models, we stitched the images in the image-set together to get object features and positions. For LXMERT, we used the same amount of layers, embedding dimensions as have been used in LXMERT and a learning rate of $2e-5$ and trained our model for 4 epochs. Adam optimizer with linear-decayed learning-rate schedule was used for this task. For VisualBERT, all the hyperparameters utilized for training the model were kept the same as in the default implementation provided by the authors.

For all the baseline models we use the Categorical Cross-Entropy loss for training. 

\section{Proposed Approach}
\label{sec:proposed-approach}

\begin{figure}[!ht]
\begin{center}
\includegraphics[width=3in]{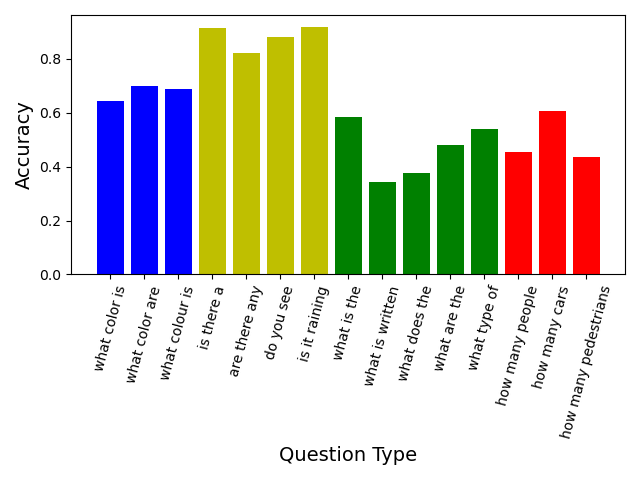}
\caption{Question-based error analysis for LXMERT baseline}
\label{fig:ques-anal}
\end{center}
\vskip -0.2in
\end{figure}

An analysis of the relationship between the model performance and the nature of questions shows that certain question categories such as count-based and colour-based questions perform worse as compared to other question categories such as object-based and true/false questions. This is also shown in Fig. \ref{fig:ques-anal} where we analyze the 15 most frequent types of questions in our dataset using LXMERT. This can be attributed to the inherent difficult nature of these tasks, often involving a multi-step process; including object detection, recognition and counting; to perform optimally while also limiting the impact of common issues such as double-counting. 

We propose models using pre-training (Section \ref{pre-train}) and count-aware feature generation (Section \ref{count-graph} and \ref{count-loss}) for mitigating these issues. Further, the questions in our dataset have been susceptible to dataset biases. We propose using Adversarial regularization to remove language dependence in LXMERT in Section \ref{adv-reg}.

\subsection{Removing Language Dependence using Adversarial Regularization}
\label{adv-reg}

Adversarial regularization is the process of marginalizing biases which may be detrimental to performance of a model when evaluated on out of distribution data. The procedure of performing regularization adversarially involves introducing a competing framework which penalize overconfident predictions of the model when data with insufficient information is provided to the model.

The idea to incorporate adversarial regularization stems from the observations made on the Question only Analysis. A model which captures strong language biases limits its ability to generalize to data which is not observed during training and hence introducing a regularizing component can prove beneficial to improve the performance of the ISVQA baselines discussed in section \ref{sec:baselines}. A similar observation to the one made in the Question only BERT baseline is observed in the LXMERT model. Here the LXMERT model which achieves the best result for ISVQA is able to accurately answer questions even when the information from the visual modality is completely removed by zeroing out object features extracted from the RCNN module. The Table \ref{tab:advreg} shows the result from our experiment on adversarial regularization.

The procedure for adversarial regularization involves eliminating features of objects in the image that contain the answer to the question. This is done by parsing the question to identify objects (ex. car, person, truck, etc.) and removing their image feature representation information extracted by the Faster-RCNN module. The training example generated by the above process lacks important information needed to answer the question and hence the model must not be able to predict the correct answer for such an example. Correct answers to such questions might indicate the presence of unwanted biases learnt by the model during training and to eliminate them we reverse the direction of gradient on adversarial examples and add an additional term to the original loss formulation which maximizes the loss on adversarial examples, thus eliminating biases in the process. The equation \ref{eq:adversarial} shows the formulation of the loss function with an additional loss term for adversarial examples.

\begin{equation*}
\label{eq:adversarial}
    L_(q, I, I'; \theta) = L_(q, I; \theta) - \lambda_{R} * L_(q, I'; \theta)
\end{equation*} 

Here, $L$ corresponds to the Loss of the ISVQA model. $\lambda_{R}$ is the regularization constant and controls the strength of the regularization. $q$ represents the question, $I$ represents the true image features extracted from Faster-RCNN, $I'$ represents the augmented Image features and $\theta$ represents the model parameters. For all our experiments on adversarial regularization we fix $\lambda_{R}$ to be $0.1$.

We also experiment with 2 different loss functions for adversarial regularization and analyze their effectiveness in reducing bias. The Cross-Entropy (CE) loss obtained on an adversarial example given the ground truth label gives us a measure of the extent to which the model is overconfident in its prediction hence we choose to maximize this loss. Another loss function that we try out is the per class Binary Cross-Entropy (BCE) loss which is used to determine the cumulative loss per class given an adversarial example and we choose to minimize this loss as its magnitude indicates how certain the model is of the predicted answer. Hence a higher value indicates more certainty which corresponds to overconfident predictions.

\subsection{Enhancing Pre-Training with Color and Position Questions}
\label{pre-train}
Fig.~\ref{fig:ques-anal} shows that color based questions perform sub-optimally for ISVQA. We try to tackle this, along with the general problem of spatial-awareness (which is pivotal to ISVQA) using enhanced pre-training. Recent works in the field of NLP have focussed on different positional embedding strategies to improve performance with Transformer-based models. While pre-training using novel positional embedding strategies would have been an interesting path to pursue, the computational complexity of the problem would have been difficult to manage within the resources available (LXMERT pre-training took 10 days on 4 Titan Xp). In order to achieve the benefits of having a better spatial and color-awareness, we take advantage of the pre-training strategy used by LXMERT. In particular, we take advantage of the fact that LXMERT's last 10 epochs of pre-training only take place on the Image Question-Answering task. In order to provide better spatial and color information, we continue pre-training the previously pre-trained LXMERT on artificially created color and position based VQA questions. 

\begin{figure}[!ht]
\begin{center}
\includegraphics[width=3in]{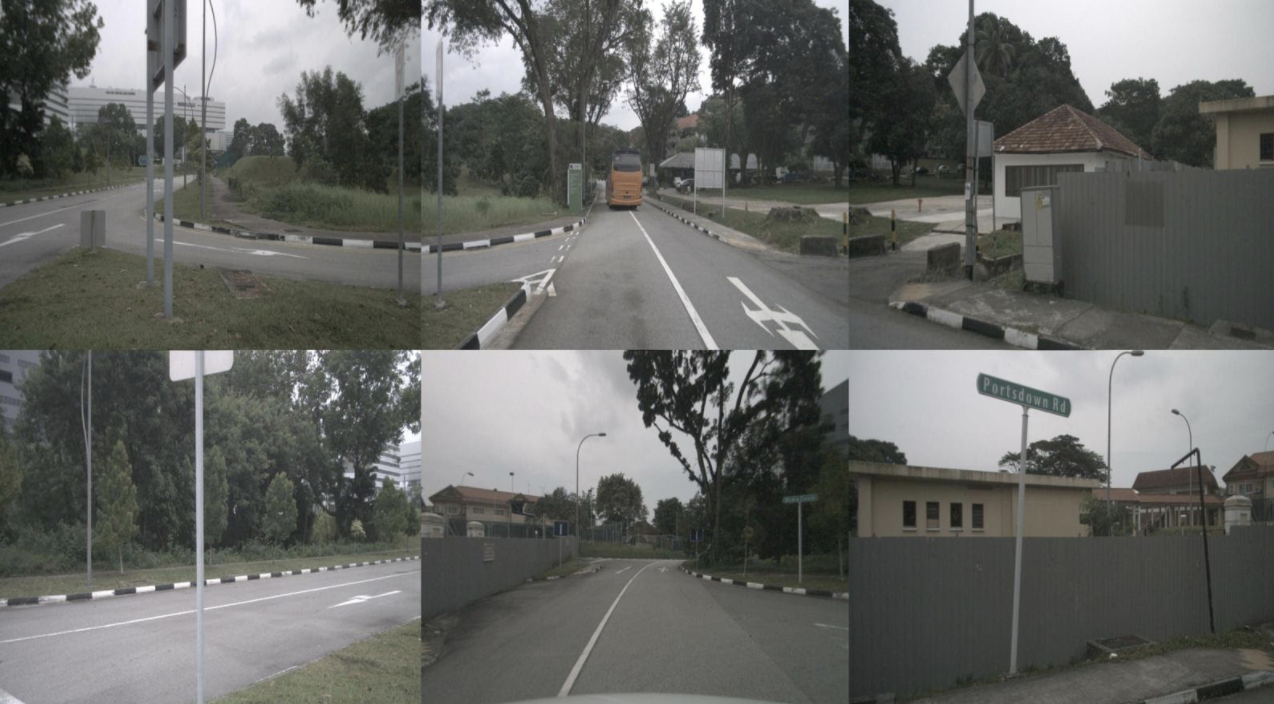}
\caption{The given image generates two questions: \textit{What is the color of the sign?} (Green) and 
\textit{What is below the wall?} (Sidewalk)}
\label{fig:pre-train-img}
\end{center}
\vskip -0.2in
\end{figure}

In order to create artificial color VQA questions, we used the attributes predicted by R-CNN. For all the color-based attributes, we created a template question \textit{What is the color of X} where X is the object. For generating position questions, we looked into the left, right, top, and bottom-most objects, found the nearest objects to these objects based on Euclidean distance, and generated artificial question of the form \textit{what is below X}. We only consider the left-most object as the answer for a question of the format \textit{what is to the left of X} to ensure that there is no ambiguity. Further, we also ensure that the questions are generated for unique objects - i.e. objects that appear in the questions only appear once in the set of the images. Fig.~\ref{fig:pre-train-img} gives an example of questions generated for an image set using this technique. 

For this apporach, we use the same hyperparameter, loss, and optimizer as with the LXMERT baseline.

\subsection{Count Aware ISVQA using graph based deduplication}
\label{count-graph}

We observe that all the baseline models struggle to perform optimally on the count-based questions. Traditionally, these questions have been susceptible to dataset biases \cite{jabri2016revisiting} and often require a multi-step process involving object detection, recognition and counting to perform optimally, while also maintaining the generalizability of the model. We propose an  improvement on the baseline model by adapting and extending the work of \cite{zhang2018learning} specifically to handle count based questions.
In their work \cite{zhang2018learning} propose that simply using soft attention for the VQA task does not help with the objective of counting. The major reason for this is attributed to the fact that the soft attention mechanism gives an equal weight to an object category. Thus if $n$ objects of the same class are present in an image the attention weights for each are $1/n$ times reduced (with all of them still summing to 1). 

The authors present a differentiable mechanism for counting from attention weights, while
also dealing with the problem of overlapping object regions of interest to reduce double-counting of objects.
The central idea is to create a graph and subsequently an adjacency matrix from these object proposals. Edges of the graph can then be scaled and pruned in a specific way such that a count of the number of underlying objects is found. The model takes in input as the region of interest features and the corresponding bounding boxes extracted from a Faster R-CNN network and the question encoded using an LSTM network.
Adjacency graph $A$ is obtained from the attention matrix by computing the outer product - $A = aa^T$. In this graph, the $i^{th}$ vertex represents the object region associated with $a_i$ and the edge between any pair of vertices $(i, j)$ has weight $a_i a_j$. In the regions associated with bounding boxes, the matrix $A$ forms the basis for the graph between objects. Further pruning is done in two steps using an inter edge pruning and an intra edge pruning technique.

Bounding boxes are compared using, intersection-over-union (IoU) metric. Hence the distance matrix $D$ is defined as $D_{ij} = 1 - IoU(b_i , b_j )$. 
Intra object edges are removed by elementwise multiplying the distance matrix with the attention matrix $A^\prime = f_1(A)  f_2(D)$ where $f_1, f_2$ represent piecewise linear functions. For removing inter-object edges a similarity metric is computed over the rows of $A^\prime$. The whole process results in the generation of the score matrix $C$ which is the count based score for each of the objects. 

We adapt this approach to the task of ISVQA as an intermediate step to improve the features extracted by the RCNN and score each ROI according to the computed matrix $C$. Each of the image features is multiplied by its respective score and then fused with the question representation using soft attention. This output is considered as the updated ‘count-aware’ feature. The count aware features are used as an input to the LXMERT model. The count-aware embedding model uses the RCNN extracted features of dimension 2048 with 36 regions of interest in each image along with the corresponding bounding boxes. The questions are encoded with LSTM layers of hidden dimension 1024. The model is trained for 100 epochs using Adam optimizer using a learning rate of $1.5e-3$ with an Exponential Learning Rate scheduler. The model is trained on negative log likelihood loss. The LXMERT model is trained from scratch with AdamW optimizer using an initial learning rate of $2e-5$. We compare the prediction results of this model with an LXMERT model trained from scratch using base RCNN features and observe a visible improvement.

\subsection{Improving Performance on Count-based Questions using Regression Loss}
\label{count-loss}

To address the poor performance on count-based questions we also experiment with another approach which approaches such questions from the prospective of regression. Contrary to the approach in section \ref{count-graph} which involves processing inputs to the LXMERT model this approach introduces a new Mean Squared Error loss along with the classification loss for count based questions.
The process for training this model involves utilizing both classification and regression heads to predict the answer and at inference time the answer to the question is determined by using only the classification head.

\begin{table*}[h]
    \centering
    \begin{tabular}{c|c|c|c}
         \hline
         Model & Model Type & Train Accuracy (\%)& Test Accuracy (\%) \\ \hline \hline
 LXMERT & Baseline & 89.95 & 64.55 \\ \hline
 LXMERT without pretraining& Baseline & 79.37 & 57.05 \\ \hline
 Visual BERT  & Baseline & 74.3 & 55.55 \\ \hline
 HME - Video QA  & Baseline & 97.1 & 49.9  \\ \hline
          \hline
  LXMERT with enhanced pre-training & Ours & 88.74 & \textbf{64.93} \\ \hline
  LXMERT with Count-based VQA & Ours & 79.56 & 57.50 \\ \hline
  LXMERT with Regression Loss & Ours & 92.6 & 64.38 \\ \hline
  Adversarial Regularization  & Ours & 82.62 & 63.24\\ \hline
    \end{tabular}
    \caption{Baseline model results}
    \label{tab:result_table}

\end{table*}
\section{Results and Discussion}
\label{sec:results}
Table~\ref{tab:result_table} tabulates the results obtained for the baseline and the advanced models proposed by us on the ISVQA dataset. We further assess the performance of each research enhancement in the following sections.

\subsection{Removing Language Dependence using Adversarial Regularization (In-Depth Analysis)}

\begin{figure}[!ht]
 \begin{center}
 \includegraphics[width=3in]{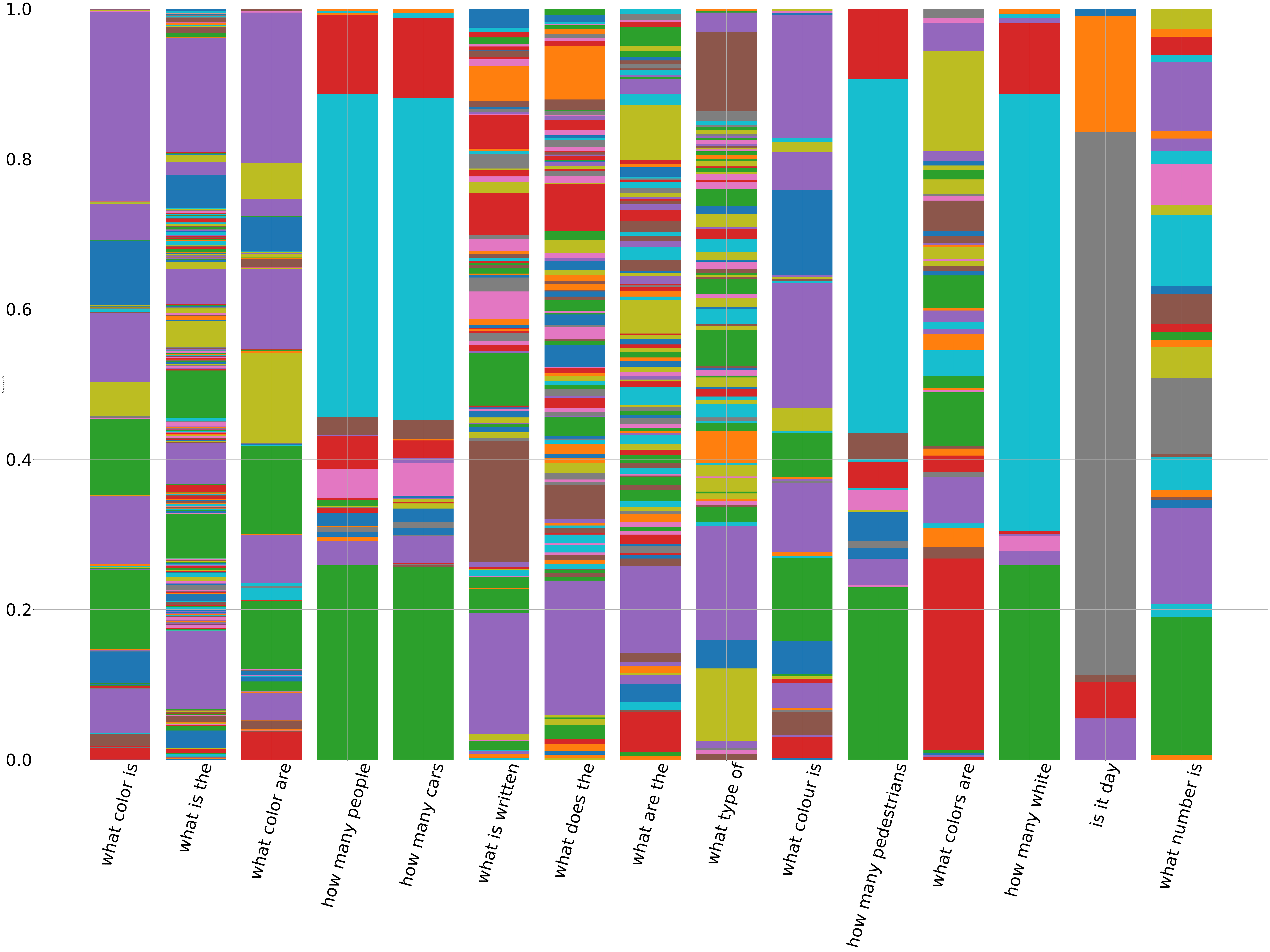}
 \caption{Frequency of answer categories for 15 most frequent questions. Each column bar corresponds to a question-type and each colored-stack in the bar refers to an answer category}
 \label{fig:word_freq}
 \end{center}
 \vskip -0.2in
 \end{figure}

To better understand the  bias present in the text dataset and its implications on the model performance, we analyze the distribution of answer categories across each questions and find that colour-based and count-based questions are often dominated by a smaller number of answer categories, as shown in Figure \ref{fig:word_freq}. We can see that for count-based questions like \textit{how many people/cars/pedestrian etc}, same 2 answer categories occurred about 75\% times (denoted by blue and green stacks), while all the remaining classes accounted for about 25\% of the occurrences, thus denoting a high answer bias for count-based questions. Meanwhile, object-based questions like \textit{what does the/what are the} show a high diversity of answer classes, and thus low text-induced bias. This indicates the presence of bias in the ISVQA dataset and can lead the models learned on this data to not generalize well and under-utilize visual information when answering questions as the information needed to answer the questions can be learnt from just the language modality features. This is further highlighted from Table \ref{tab:question-only} which shows the results for a BERT-based question-only model. We observe that this unimodal model is able to achieve over 52\% accuracy.

\begin{figure}[!ht]
 \begin{center}
 \includegraphics[width=3in]{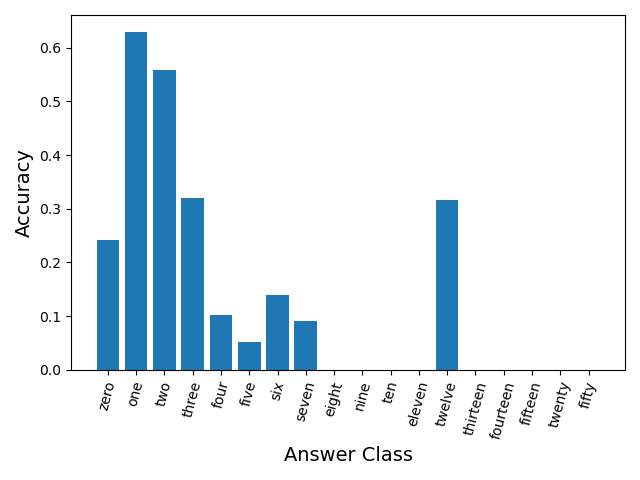}
 \caption{Accuracy of count-based answer categories}
 \label{fig:count_acc}
 \end{center}
 \vskip -0.2in
 \end{figure}

 High occurrence of certain answer categories can result in model being biased towards those answers. This is evident in Figure \ref{fig:count_acc} where we show the accuracy of count-based answer categories for the baseline LXMERT model. We can see that more frequent answer categories such as \textit{Three/Two} had accuracy as high as 55\%, while rare categories such as \textit{nine/fifteen} were never predicted correctly.

\begin{figure*}[!htb]
\begin{center}
\subfloat[Train set \label{fig:ans-freq-train}]{\includegraphics[scale=0.3]{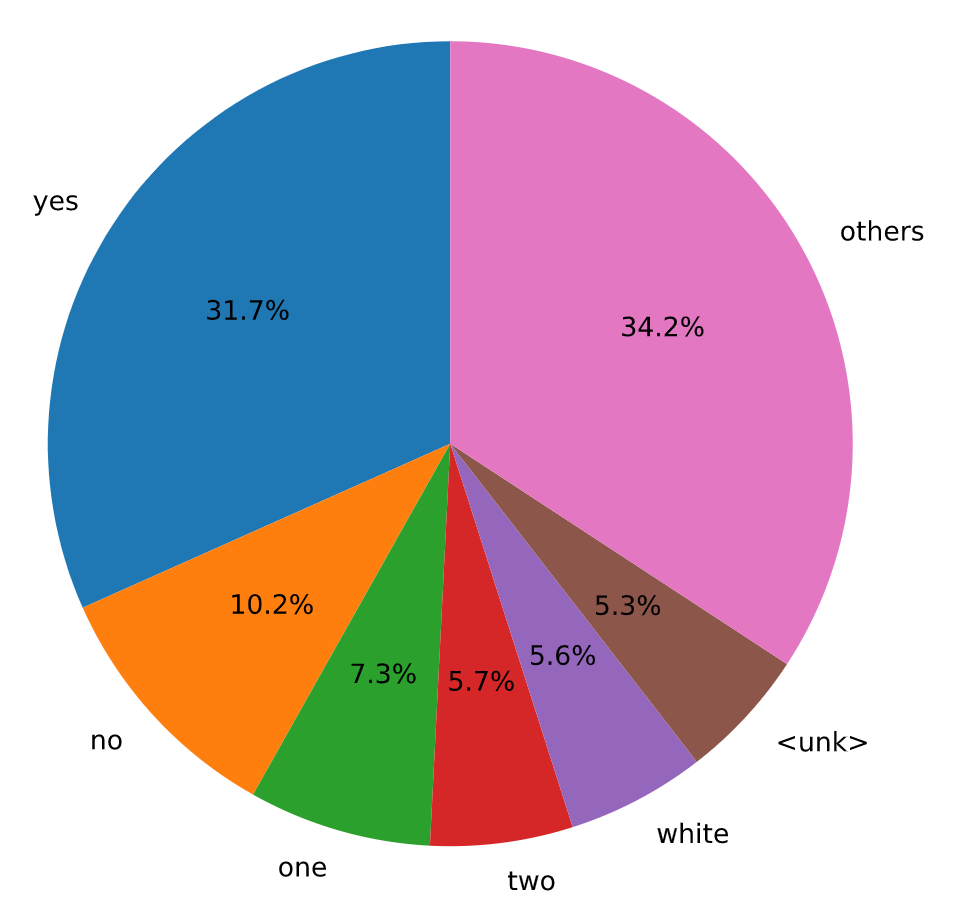}}
\hspace{0.3in}
\subfloat[Test set \label{fig:ans-freq-test}]{\includegraphics[scale=0.3]{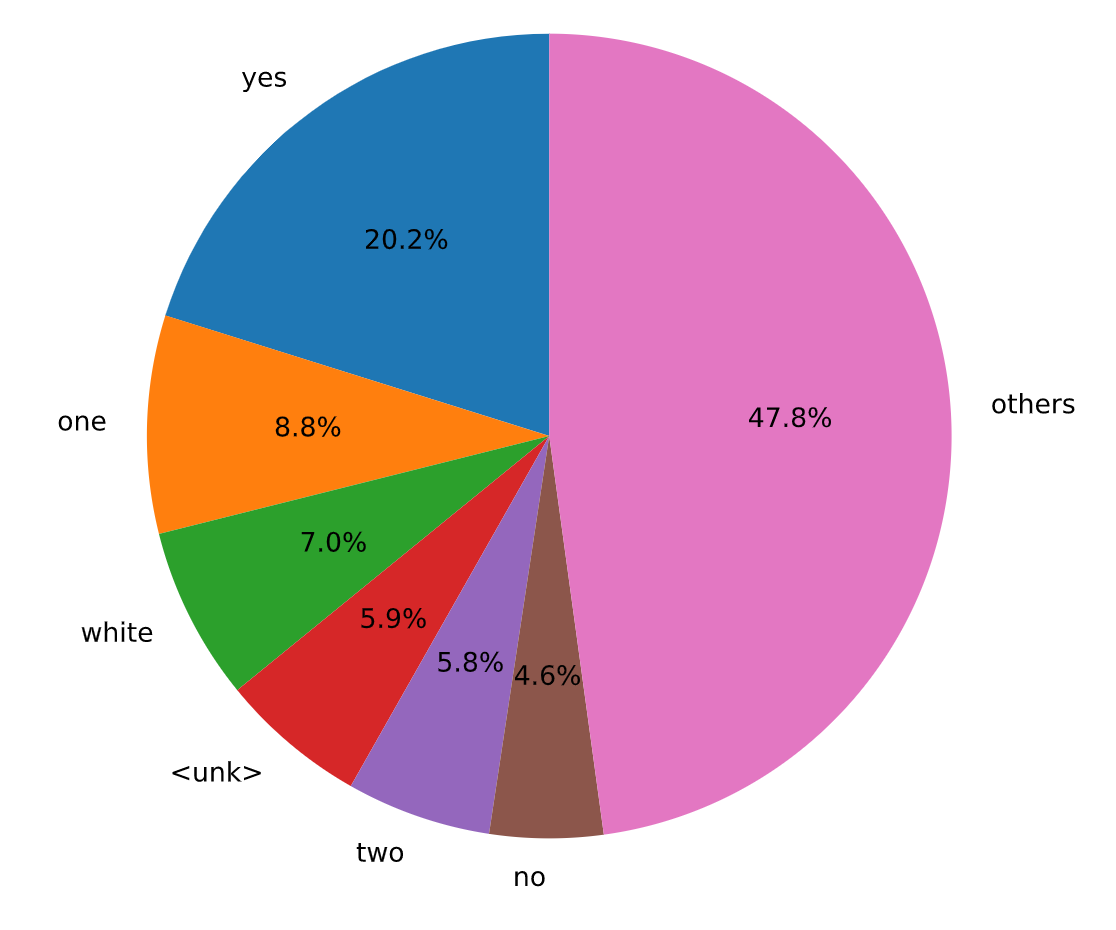}}
\caption{Most Frequent Answers}
\label{fig:ans-freq}
\end{center}
\end{figure*}

\begin{table}[!htp]
    \centering
    \begin{tabular}{c|c|c|c}
         \hline
         & Base LXMERT & w/ CE Reg. & w/ BCE Reg. \\
         \hline \hline
          L + V  & 64.55\% & 63.24\% & 63.05\%\\ 
          L & 42.87\% & 5.66\% & 12.8\% \\ 
    \end{tabular}
    \caption{Results for Adversarial Regularization. Here L corresponds to Language and V corresponds to visual modality}
    \label{tab:advreg}
\end{table}

To mitigate the presence of such biases we introduce a new adversarial regularization formulation. Here we experiment with training the LXMERT model using 2 different regularization loss terms which penalizes overconfident predictions on adversarial examples and to compare the effect of adversarial regularization we utilize the baseline LXMERT model which was trained on both Language and Visual Modalities. Our results presented in Table \ref{tab:advreg} indicate that when visual information is scrubbed (zeroed out) during inference time (but is trained with both Visual and Language modality present) the LXMERT model trained without adversarial regularization is able to obtain considerably high accuracy on the test dataset showing its over reliance on language modality and the presence of bias in the test dataset. However, when the model is trained with adversarial regularization, performance on test set with srubbed out visual information is reduced drastically ~37\% while little impact is seen when both the visual and text data are provided which indicates that regularized model does not over rely on language biases. We also observe that using the label-wise Binary Cross-Entropy loss is less effective in removing bias when compared to using the Cross-Entropy adversarial loss formulation.

\begin{table}[!htp]
    \centering
    \begin{tabular}{p{3cm}|p{3cm}}
         \hline
          Train Accuracy (\%)& Test Accuracy (\%) \\ \hline \hline
72.55 & 52.69 \\ \hline
    \end{tabular}
    \caption{Question-only BERT result}
    \label{tab:question-only}
    
\end{table}

\subsection{Enhancing Pre-Training with Color and Position Questions}
Table~\ref{tab:result_table} shows that our LXMERT model with enhanced pre-training performs slightly better than the baseline LXMERT model. This shows the efficacy of pre-training on our artificially created data. Further, from Table~\ref{tab:color-res}, we see that our performance on the color-based questions is significantly better than the baseline LXMERT model. We do not report results on position-specific questions as the aggregator for such questions is not provided in the dataset. These results re-iterate the need to provide better color representations to our model as they can significantly enhance the performance. More importantly, these results provide motivation on developing and spending computational resources on training better pre-training objectives that incorporate position and color as they can help improve accuracy on ISVQA even further.

While using such artificially generated questions leads a decent performance improvement for Color based questions, there is a clear disadvantage of using this approach as the errors in object or attribute prediction from R-CNN get cascaded to the questions generated, making our model susceptible to these errors. An of example of this is the question \textit{what is the color of the train} with answer \textit{orange} for image in Fig.~\ref{fig:pre-train-img}. In this case, our R-CNN mis-detects the bus as a train and leads to the creation of an incorrect question.

\begin{table}[!htp]
    \centering
     \begin{tabular}{c|c}

         \hline
         Model & Color-Only Accuracy\\
         \hline
          LXMERT & 67.72 \\ 
          LXMERT w/ enhanced PT
& \textbf{70.09} \\ 
    \end{tabular}
    \caption{Model performance on Color-based Questions only}
    \label{tab:color-res}
\end{table}

\subsection{Count Aware ISVQA using graph based deduplication}
The results for the Count based ISVQA is presented in table  \ref{tab:result_table}. The results are obtained on an LXMERT model trained from scratch rather than adopting a pretrained one. Thus the comparison will be done on an LXMERT model without pretraining using base RCNN features. This is done because the input features for the Count based model are changed and thus we cannot adapt a pretrained model as is (and pre-training our model is computationally unfeasible). A slight improvement over the baseline is observed on the overall results whereas Table \ref{tab:count-graph} compares the model performance specifically on count based questions. This shows a clear improvement over the baseline confirming that 'count-aware' features perform better. The claims made by \cite{zhang2018learning} that the soft attention based models generally do not perform well on count based question is proven. This also shows that a graph based modelling of the regions of interest in an image is a good approach to resolve disambiguities and solve the double counting problem especially in datasets such ISVQA where such anomalies are present in abundance. 

\begin{table}[!htp]
    \centering
    \begin{tabular}{c|c}
         \hline
         Model & Count-Only Accuracy\\
         \hline
          LXMERT w/o pre-training & 42.7\% \\ 
          LXMERT w/ Count Based VQA
& 46.5\% \\ 
    \end{tabular}
    \caption{Model performance on Count-based Questions only}
    \label{tab:count-graph}
\end{table}

\subsection{ Improving Performance on Count-based Questions using Regression Loss}
The results for this experiment in Table \ref{tab:result_table} and \ref{tab:count-reg-res} show similar results to the LXMERT baseline and a marginal 0.2\% improvement in Count-only question accuracy was observed. The results suggest that introducing an additional regression loss is enough to improve accuracy on count-based questions. And  additional modifications such as changes in the features extraction process, better attention mechanisms to improve alignment and overall enhancements to the model architecture are required to improve accuracy on count based questions.

\begin{table}[!htp]
    \centering
    \begin{tabular}{c|c}
         \hline
         Model & Count-Only Accuracy\\
         \hline
          LXMERT & 49.84\% \\ 
          LXMERT w/ Regression Loss
& 50.02\% \\ 
    \end{tabular}
    \caption{Model performance on Count-based Questions only}
    \label{tab:count-reg-res}
\end{table}

\section{Conclusion and Future Work}
\label{sec:conclusion}
In this work we discuss 4 research ideas to improve performance on Image Set Visual Questions Answering and present an in depth analysis for the problem of Language Dependence along with testing 2 different adversarial regularization formalizations. We utilize three established baselines for the task and perform error analysis to provide a strong motivation for our research ideas. Our presented approaches include an enhanced pertaining strategy for the task of ISVQA that incorporates color and spatiality. We then describe ideas to incorporate count awareness in deep learning models for ISVQA. Finally, we discuss an adversarial regularization technique for bias removal technique as we observe the presence of significant bias in both training and testing data. We observe that our proposed approaches are able to improve performance over the LXMERT baseline, with a 0.38\% absolute performance increase using our enhanced pre-training strategy. Our count-based models were able to improve performance on count-based questions by 3.8\% while enhanced pre-training helped improved the performance on color-based questions by 2.3\%. Additionally, our adversarial regularization technique is able to reduce dependence on the language modality.

As a future work, we suggest the development of a ISVQA dataset with changing priors similar to VQA-CP to study the generalizability of ISVQA models. Furthermore, work needs to be done to improve the performance on count based questions as our experiments reveal that count based questions are particularly harder for Visual Questions Answering models to handle. Finally, our results from the enhanced pre-training model show that incorporating more pre-training objectives that cater to spatial and color-awareness can help improve the performance even further and running such experiments can be worth the computational efforts.

\bibliography{example_paper}
\bibliographystyle{icml2020}

\newpage


\end{document}